%% file: main_supp.tex
\documentclass[10pt,twocolumn,letterpaper]{article}

\usepackage{wacv}                       % To produce the CAMERA-READY version

\definecolor{wacvblue}{rgb}{0.21,0.49,0.74}
\usepackage[pagebackref,breaklinks,colorlinks,allcolors=wacvblue]{hyperref}

\input{preamble}

\def\wacvPaperID{1940} % *** Enter the WACV Paper ID here
\def\confName{WACV}
\def\confYear{2027}

\usepackage{colortbl}
\usepackage{xcolor}
\usepackage{graphicx}
\usepackage{amsmath}
\usepackage{amssymb}
\usepackage{booktabs}
\usepackage{multirow}

\usepackage{kotex}

\newcommand{\model}{AFO}
\definecolor{ab_best}{RGB}{168, 192, 251}
\definecolor{ab_better}{RGB}{193, 214, 251}
\definecolor{ab_good}{RGB}{229, 234, 251}

\title{All-For-One: Distilling Visual Foundation Models \\ to a Fast, Memory-Efficient Visuomotor Policy}

\author{
Byungkun Lee$^{1}$\thanks{Equal contribution.} \quad
Minho Park$^{1}$\footnotemark[1] \quad
 Hojoon Lee$^{2}$  \quad 
Dongyoon Hwang$^{1}$ \\ Donghu Kim$^{2}$ \quad Hyunseung Kim$^{1,3}$ \quad 
Kyungmin Lee$^{1}$ \quad Yunho Kim$^{4}$ \quad Jaegul Choo$^{1}$ \\
$^{1}$KAIST \quad $^{2}$Holiday Robotics \quad
$^{3}$KRAFTON \quad $^{4}$Neuromeka\\
{\tt\small \{byungkun.lee, m.park\}@kaist.ac.kr}
}

\usepackage{listings}
\usepackage{docmute}
\begin{document}

% LaTeX clears title commands after the first \maketitle; save the originals.
\makeatletter
\let\combined@maketitle\maketitle
\let\combined@internalmaketitle\@maketitle
\makeatother

% The main document prints the shared References before the supplement.
\input{main}

\clearpage

% Start the unmodified supplementary document on a fresh page.
% Its own section/table/figure formatting is kept; pages remain continuous.
\makeatletter
\let\maketitle\combined@maketitle
\let\@maketitle\combined@internalmaketitle
\let\title\titleold
\gdef\@author{}
\gdef\@thanks{}
\makeatother
\setcounter{equation}{0}
\setcounter{footnote}{0}
% Distinct hyperlink destinations for the counters reset by the supplement.
\renewcommand{\theHsection}{supp.\arabic{section}}
\renewcommand{\theHtable}{supp.\arabic{table}}
\renewcommand{\theHfigure}{supp.\arabic{figure}}
\renewcommand{\theHequation}{supp.\arabic{equation}}

% Keep every supplementary citation, but suppress its duplicate References.
% With \input, all citations are written to the same .aux file, including
% works cited only in the supplement; BibTeX includes them in the list above.
\begingroup
\renewcommand{\bibliographystyle}[1]{}
\renewcommand{\bibliography}[1]{}

\input{supplementary}

\endgroup

\end{document}

%% file: preamble.tex
\usepackage[accsupp]{axessibility}

\usepackage{pifont}% http://ctan.org/pkg/pifont
\newcommand{\cmark}{\ding{51}}%

\usepackage{array}
\newcolumntype{P}[1]{>{\centering\arraybackslash}p{#1}}
\newcolumntype{M}[1]{>{\centering\arraybackslash}m{#1}}
\newcolumntype{L}[1]{>{\hspace{0.5em}\raggedright\arraybackslash}m{#1}}
\newcolumntype{R}[1]{>{\raggedleft\arraybackslash}m{#1}}

\usepackage[capitalize]{cleveref}
\Crefformat{figure}{Fig.~#2#1#3}
\Crefformat{equation}{(#2#1#3)}
\Crefformat{table}{Tab.~#2#1#3}

\usepackage{tabularx,booktabs}
\newcolumntype{Y}{>{\centering\arraybackslash}X}
\usepackage{colortbl}
\usepackage{multirow}
\usepackage{soul}
\usepackage{wrapfig}
\newcommand{\drule}{\specialrule{0.2pt}{1pt}{1pt}%
\specialrule{0.2pt}{0pt}{\belowrulesep}%
}
\newcommand{\std}[1]{{\small$\pm$#1}}

\usepackage{titletoc}

%% file: main.tex
\maketitle

%%%%%%%%%%%%%%%%%%%%%%%%%%%%%%%%%%%%%%%%%%%%%%%%%%%%%%%%%%%%%%%%%%%%%%%%%%%%%%%%
\begin{abstract}

\vspace{-0.15cm}
% VFM
% - SAM, DAM, Molmo
% - Theia, DINO, MVP, R3M, ...
% 
% why small cnn is better than VFMs?

Visual Foundation Models (VFMs) provide semantic cues such as depth and segmentation useful for robotic manipulation, but directly deploying them in closed-loop control is impractical due to high computational cost, temporal inconsistency, and distribution shift between static internet images and egocentric robot observations.
We propose All-For-One (AFO), a two-stage distillation framework that leverages VFMs strictly as privileged information during training, using downstream robotic data. 
Following the privileged learning paradigm, AFO decouples semantic representation learning from inference-time policy execution.
In the All stage, a teacher policy fuses RGB observations with outputs from multiple VFMs to learn a semantically enriched representation. 
In the For-One stage, this knowledge is distilled in-distribution into a lightweight RGB-only student policy. 
Evaluated on two Adroit tasks and five MetaWorld tasks in simulation, as well as three real-world manipulation tasks, AFO outperforms prior foundation-model-based approaches while using
over 30$\times$ fewer parameters than VFM-based encoders and 50$\times$ lower latency than its own teacher in simulation benchmarks,
providing an efficient pathway for transferring foundation-model knowledge to robotic control.

\end{abstract}

%%%%%%%%%%%%%%%%%%%%%%%%%%%%%%%%%%%%%%%%%%%%%%%%%%%%%%%%%%%%%%%%%%%%%%%%%%%%%%%%
\vspace{-0.2cm}
\section{Introduction}
\vspace{-0.05cm}

Robust visual perception is a prerequisite for robotic manipulation in unstructured environments. 
To interact effectively with the physical world, a robot must infer rich semantic and spatial information from visual observations, including the location of task-relevant objects, their boundaries, and depth.
Such information directly influences downstream control~\cite{desouza2002vision_nav_survey,kroemer2021review, billard2019trends, sunderhauf2018limits, bohg2014datadriven, florence2018dense}.
However, robotic learning systems typically operate under severe data constraints: collecting real-world demonstrations is expensive, time-consuming, and task-specific. 
This data scarcity has motivated extensive interest in leveraging general-purpose vision foundation models (VFMs) that can transfer across tasks and environments.

\begin{figure}[t]
\begin{center}
\includegraphics[width=1.0\columnwidth]{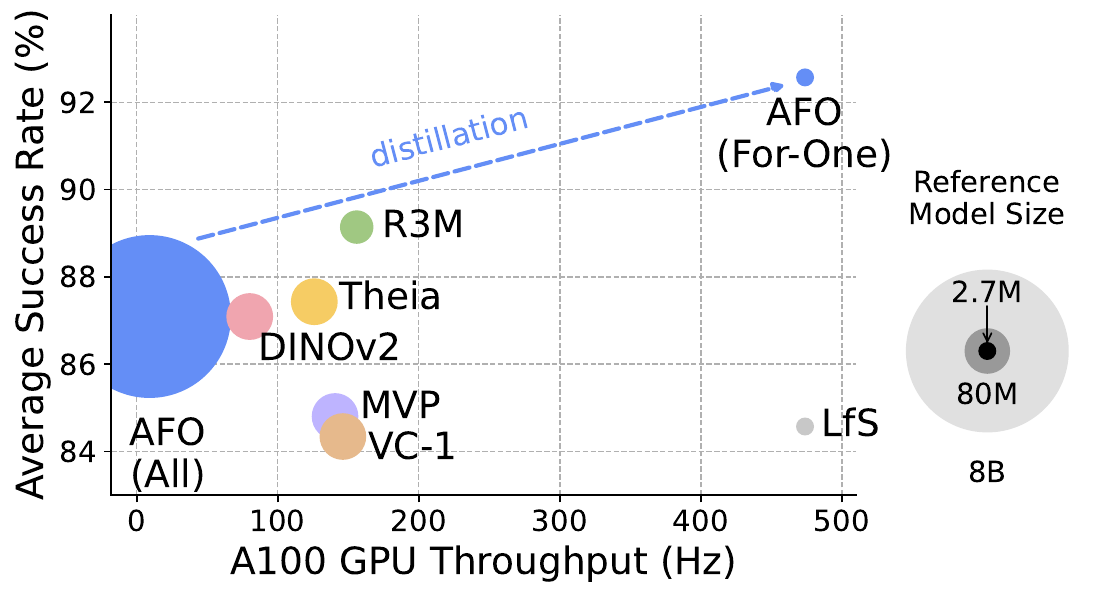}

\end{center}
\vspace{-0.5cm}
\caption{
\textbf{Overcoming the Compute--Performance Tradeoff.}
Learning from scratch (LfS)~\cite{hansen2022lfs} is fast but inaccurate, whereas VFM backbones~\cite{radosavovic2022mvp,shang2024theia,oquab2023dinov2,majumdar2023vc1,nair2022r3m} are accurate but slow.
By distilling large-scale pretrained VFMs~\cite{deitke2025molmo, ravi2024sam2, depth_anything_v2} into a compact student, AFO attains high success rates while remaining computationally efficient at inference.
}
\label{fig:fig1}
\vspace{-0.5cm}
\end{figure}

To leverage vision foundation models (VFMs), one line of work directly adopts off-the-shelf foundation encoders such as DINOv2~\cite{oquab2023dinov2} or CLIP~\cite{radford2021clip} as frozen perception backbones, or propose robot-centric pretraining~\cite{radosavovic2022mvp,nair2022r3m,majumdar2023vc1} and distillation methods~\cite{shang2024theia} from large-scale data~\cite{ImageNet,grauman2022ego4d}.
However, such pretrained encoders are typically large~\cite{he2016resnet,dosovitskiy2020image}, incurring an unfavorable compute--performance tradeoff (\Cref{fig:fig1}).
Moreover, since they are often pre-trained on general web-crawled images (\eg, ImageNet~\cite{ImageNet}), a substantial distribution shift arises between static, exocentric web imagery and egocentric, task-driven robot observations.
As a result, these representations usually yield only marginal gains over small models trained from scratch, or even degrade performance~\cite{hansen2022lfs,majumdar2023vc1}.

Recently, VFMs have advanced beyond generic vision encoders.
For example, Molmo~\cite{deitke2025molmo} localizes objects via pointing from free-form natural language, SAM~\cite{kirillov2023sam,ravi2024sam2,carion2025sam3} produces fine-grained segmentation masks from a point prompt, and DAM~\cite{depth_anything_v1,depth_anything_v2,depth_anything_v3} predicts reasonable depth maps from RGB images alone.
In principle, these capabilities could help overcome the data limitations of robotic learning.
In practice, however, their tremendous computational and memory footprints preclude real-time, on-device inference in closed-loop control, rendering their predictions available only as privileged signals at training time.
In this paper, we explore a distillation method that transfers this privileged knowledge from VFMs into compact models suitable for real-time control.

To this end, we propose \textbf{All-For-One (AFO)}, a two-stage distillation framework that leverages visual foundation models (VFMs) strictly as privileged information during training. 
In the \textbf{All} stage, a teacher policy is trained on downstream robotic data, fusing RGB observations with depth and segmentation outputs from multiple large-scale VFMs to learn a semantically enriched representation. 
In the \textbf{For-One} stage, this VFMs' privileged knowledge is distilled into a lightweight, RGB-only student policy using only robotic data.
The student is trained end-to-end from raw RGB observations by imitating expert actions while aligning its internal representation with the teacher’s VFM-derived features, thereby transferring privileged visual knowledge without requiring VFMs at inference time.

We evaluate AFO on seven tasks across two simulation benchmarks~\cite{rajeswaran2018adroit, yu2020metaworld}, as well as three real-world manipulation tasks on a 6-DoF Indy7 robotic arm.
As shown in \Cref{fig:fig1}, AFO successfully transfers privileged VFM knowledge into a compact RGB-only policy.
Notably, the end-to-end trained student surpasses not only the baselines but also its heavy VFM-based teacher, as it adapts its representation to the target domain while bypassing the VFMs' occasional prediction errors that we analyze in detail.
Overall, AFO offers a practical and efficient pathway for transferring foundation knowledge to robotic policies while preserving real-world deployability.
Our contributions are threefold:
\begin{enumerate}[leftmargin=1.5em]
    \item We propose \textbf{All-For-One (AFO)}, a two-stage framework that treats multiple VFMs as privileged information at training time and distills their depth and segmentation knowledge into a lightweight, RGB-only policy that requires no VFMs at inference.
    \item We show that the end-to-end distilled student not only matches but surpasses its VFM-based teacher, by adapting VFM-derived representations to the target domain while avoiding the teacher's reliance on error-prone, costly VFM inference in the loop.
    \item We validate AFO on seven simulated tasks across two benchmarks and three real-world tasks on a 6-DoF Indy7 arm, demonstrating strong performance and robustness to visual distribution shifts at a fraction of the teacher's computational cost.
\end{enumerate}

\begin{figure*}[t]
\begin{center}
\includegraphics[width=0.96\textwidth]{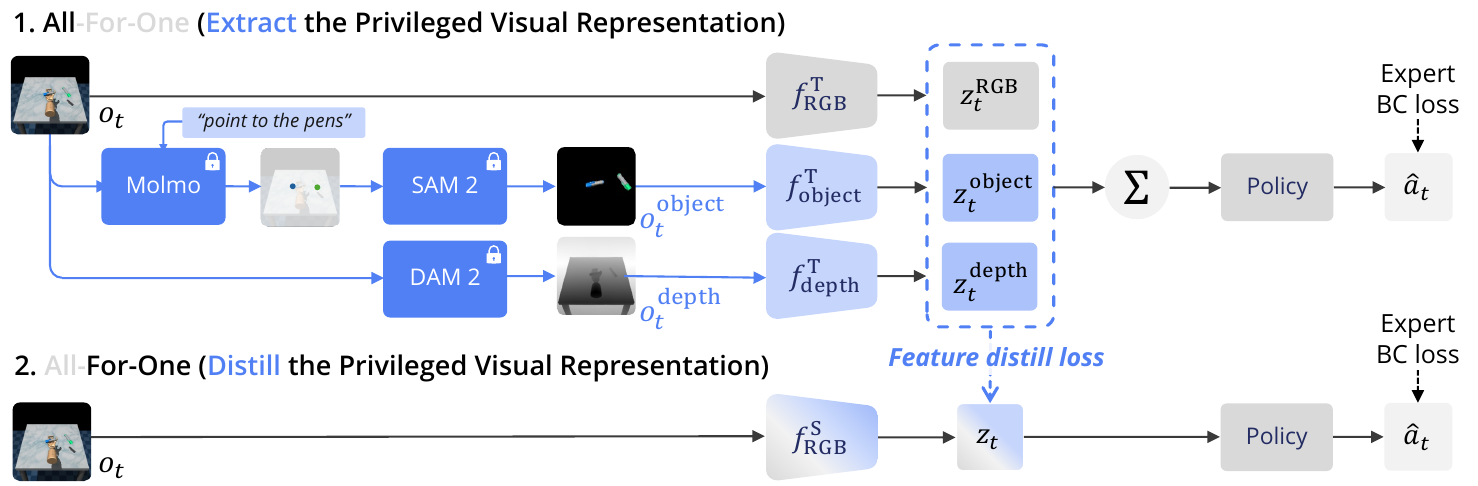}
\end{center}
\vspace{-0.4cm}
\caption{
\textbf{Overview of the All-For-One (AFO) Framework.} 
\textbf{(a) All.} A teacher network employs multiple VFMs (\eg, Molmo~\cite{deitke2025molmo}, SAM 2~\cite{ravi2024sam2}, and DAM 2~\cite{depth_anything_v2}) with learnable lightweight encoders 
 ($f^\text{T}_\text{RGB}$, $f^\text{T}_\text{object}$, $f^\text{T}_\text{depth}$).
The resulting representations are aggregated and used to train a policy with a BC loss.
\textbf{(b) For One.} A student lightweight RGB encoder 
($f^\text{S}_\text{RGB}$)
solely learns from the teacher network via a feature distillation loss with BC loss, which benefits from the rich knowledge of large VFMs.
}
\vspace{-0.5cm}
\label{fig:main pipeline}
\end{figure*}

\section{Related Work}
\label{section: related work}

\subsection{Visual Representations for  Manipulation}

\paragraph{Visual pretraining for manipulation.}
The efficacy of visuomotor control relies heavily on the quality of the underlying visual representations. Early approaches demonstrated that standard ResNet-based encoders, trained via self-supervised contrastive learning~\cite{parisi2022unsurprising} or temporal objectives~\cite{nair2022r3m}, could effectively support behavior cloning. 
Subsequent works scaled these paradigms with large-scale visual pre-training on human-video and robot-interaction data~\cite{majumdar2023vc1,radosavovic2022mvp,karamcheti2023voltron,zeng2024mpi,jiang2024mcr}.
While these encoders improve visual representations for manipulation, they are not explicitly designed to provide dense semantic and geometric cues, such as object masks or depth, that can be useful for fine-grained control~\cite{burns2023whatmakes}.

\vspace{-0.3cm}
\paragraph{Leveraging vision foundation models.}
The emergence of general-purpose Vision Foundation Models (VFMs), such as DINOv2~\cite{oquab2023dinov2}, Segment Anything (SAM) series~\cite{ravi2024sam2, carion2025sam3}, and Depth Anything (DAM) series~\cite{depth_anything_v2, depth_anything_v3}, has shifted the focus toward leveraging richer visual modalities. These models provide dense semantic and geometric cues that are highly valuable for robotics. 
% Recent methods have attempted to utilize VFMs either as frozen backbones~\cite{kim24openvla, kim2024manipgpt, wei2024finetuneVFM} or by adapting them via parameter-efficient fine-tuning~\cite{hwang2024coin, sharma2023roboadapter}.
Recent methods leverage pretrained VFMs for robotic manipulation ~\cite{kim24openvla, kim2024manipgpt}, while others explore parameter-efficient adaptation~\cite{hwang2024coin, sharma2023roboadapter}.
However, direct deployment of these large-scale models introduces severe latency bottlenecks~\cite{zhang2024same}.

To address this efficiency issue of VFMs, recent works have explored feature distillation. Theia~\cite{shang2024theia} and VER~\cite{wang2025ver} distill knowledge from multiple foundation models into a unified visual representation. However, these methods distill intermediate VFM features on generic image data such as ImageNet. Such representations are not explicitly optimized for downstream manipulation and may retain background or distractor information less relevant to the task. In contrast, our approach distills VFM outputs (e.g., segmentation masks and depth maps) directly on in-domain robotic data, compressing task-relevant knowledge into a lightweight RGB-only student policy (\Cref{tab:comparison_theia}).

\subsection{Privileged Learning and Distillation}

Privileged learning is a paradigm where an agent is trained with access to ``privileged'' information that is unavailable at test time. This framework has been extensively applied in reinforcement learning, particularly for bridging the gap between simulation and reality. Standard approaches, such as asymmetric actor-critic methods~\cite{Hu2024ICLR_Privileged_Sensing_Scaffolds, openai2018dex, pinto2017asymmetric}, provide the critic with ground-truth simulator states (e.g., contact forces, friction coefficients, or object poses) while limiting the actor to proprioceptive or visual sensors. Similarly, teacher-student frameworks often use algorithms like DAGGER~\cite{ross2011reduction} to distill a policy trained on perfect state information into a sensor-based student~\cite{lee2020learning, kumar2021rmarapidmotoradaptation, lee2024learning, singh2024dextrahrgb}.

Our work extends the concept of privileged learning from physical states to visual modalities. Instead of treating ground-truth physics as the privileged information, we treat the outputs of computationally expensive VFMs (depth and segmentation) as the privilege. While recent works have explored distilling specific visual features, such as image-to-LiDAR distillation~\cite{Mahmoud2024distillLidar}, multi-encoder distillation~\cite{ranzinger2024radio, Cao_2025_CVPR_movekd}, or agglomerating heterogeneous 2D and 3D teachers into one encoder~\cite{sariyildiz2025dune}, our framework is distinct in its ``All-For-One'' design. We combine diverse semantic outputs from multiple VFMs into a teacher and distill this fused knowledge into a lightweight, RGB-only student, simultaneously resolving the issues of latency and temporal instability.

\section{Proposed Method: All-For-One (AFO)}
\label{section: method}

We formulate the visuomotor control problem as learning a policy $\pi$ that maps observations to actions. At each timestep $t$, the robot receives an RGB image $o_t \in \mathbb{R}^{H \times W \times 3}$ and proprioceptive states $p_t$ (\eg, joint positions), aiming to predict the optimal action $a_t$.
Our objective is to learn a policy that achieves the high performance of Vision Foundation Models (VFMs) without incurring their prohibitive computational costs or temporal instability at inference time.
To achieve this, we propose \textbf{All-For-One (AFO)}, a privileged learning framework comprising two stages:
(1) The \textbf{All} stage (Sec.~\ref{subsection: teacher training}), where a teacher policy is trained using privileged depth and segmentation cues fused from multiple VFMs; and
(2) The \textbf{For-One} stage (Sec.~\ref{subsection: student training}), where this multi-modal knowledge is distilled into a lightweight student policy operating solely on RGB frames.

\vspace{1mm}
\subsection{All: Extract Privileged Visual Representation}
\label{subsection: teacher training}

The teacher network serves as a ``privileged agent,'' accessing rich visual modalities that are unavailable or too costly to compute during real-time deployment. We describe the acquisition of these privileged features, followed by the teacher's fusion architecture.

\vspace{-0.3cm}
\paragraph{Acquisition of Privileged Modalities.}
We utilize off-the-shelf VFMs to extract semantic and geometric information from the raw RGB input $o_t$. These models are kept frozen during training.

\textit{Object Segmentation ($o_t^{\mathrm{object}}$).}
To isolate task-relevant objects, we employ Segment Anything 2 (SAM~2)~\cite{ravi2024sam2}. Since SAM~2 requires spatial prompts, we automate this using Molmo~\cite{deitke2025molmo}, a Vision-Language Model. Given the initial frame $o_1$ and a natural language task description (e.g., ``pick up the red mug''), Molmo predicts the target point coordinates. These points prompt SAM~2 to generate the initial mask. For subsequent frames, we rely on SAM~2's efficient video tracking capabilities to propagate the mask temporally:
\begin{equation}
o_{t}^{\mathrm{object}} =
\begin{cases}
  \mathrm{SAM~2} \big( \mathrm{Molmo}(o_1, \text{prompt}) \big), & \text{if } t = 1 \\
  \mathrm{SAM~2} \big( o_t, o_{t-1}^{\mathrm{object}} \big), & \text{otherwise.}
\end{cases}
\label{eq:input_sam}
\end{equation}

\textit{Depth Estimation ($o_t^{\mathrm{depth}}$).}
To capture scene geometry, we employ Depth Anything V2 (DAM~2)~\cite{depth_anything_v2}. For every RGB frame, DAM~2 predicts a dense monocular depth map:
\begin{equation}
o_{t}^{\mathrm{depth}} = \mathrm{DAM~2}(o_t).
\label{eq:input_dam}
\end{equation}

\vspace{-0.3cm}
\paragraph{Teacher Architecture.}
As shown in \Cref{fig:main pipeline} (top), the teacher employs a multi-encoder design. We process the raw RGB ($o_t$), depth map ($o_t^{\mathrm{depth}}$), and segmentation mask ($o_t^{\mathrm{object}}$) using three separate, learnable encoders $f$  to project them into a shared latent space:
\begin{equation}
\begin{aligned}
z_t^{\mathrm{RGB}}    &= f^{\mathrm{T}}_{\mathrm{RGB}}(o_t), \\
z_t^{\mathrm{object}} &= f^{\mathrm{T}}_{\mathrm{object}}(o_t^{\mathrm{object}}), \\
z_t^{\mathrm{depth}}  &= f^{\mathrm{T}}_{\mathrm{depth}}(o_t^{\mathrm{depth}}),
\end{aligned}
\label{eq:encoders}
\end{equation}
where $z_t^m \in \mathbb{R}^{D}$ denotes the feature embedding for each modality. Each encoder maps its two-dimensional input, whether an RGB image, a segmentation mask, or a depth map, to a spatial feature map that is flattened and projected into this $D$-dimensional vector.
To synthesize a unified visual representation, we aggregate these embeddings via element-wise addition. This fused feature is concatenated with proprioceptive states $p_t$ and passed to an MLP policy head $\pi_{\theta}$:
\begin{equation}
\hat{a}_t = \pi_{\theta} \big( z_t^{\mathrm{RGB}} + z_t^{\mathrm{depth}} + z_t^{\mathrm{object}}, p_t \big).
\end{equation}
This additive fusion compels the encoders to align their respective modalities in a compatible feature space, thereby facilitating effective knowledge transfer.

\vspace{-0.3cm}
\paragraph{Teacher Objective.}
The teacher policy is trained via Behavior Cloning (BC) on the expert demonstration dataset $\mathcal{D}$. We minimize the Mean Squared Error (MSE) between the predicted action $\hat{a}_t$ and the ground truth action $a_t$:
\begin{equation}
\mathcal{L}_\text{BC} = \mathbb{E}_{(o_t, p_t, a_t) \sim \mathcal{D}} \big[ \| a_t - \hat{a}_t \|^2 \big].
\label{eq:bc_loss}
\end{equation}

\vspace{1mm}
\subsection{For-One: Distill the Privileged Representation}
\label{subsection: student training}

Directly deploying the teacher is impractical due to the massive parameter count of VFMs and the risk of temporal jitter in their outputs. To address this, we distill the teacher's knowledge into a lightweight student policy.

\begin{figure*}[t]
\vspace{5pt}
\begin{center}
\includegraphics[width=1.0\textwidth]{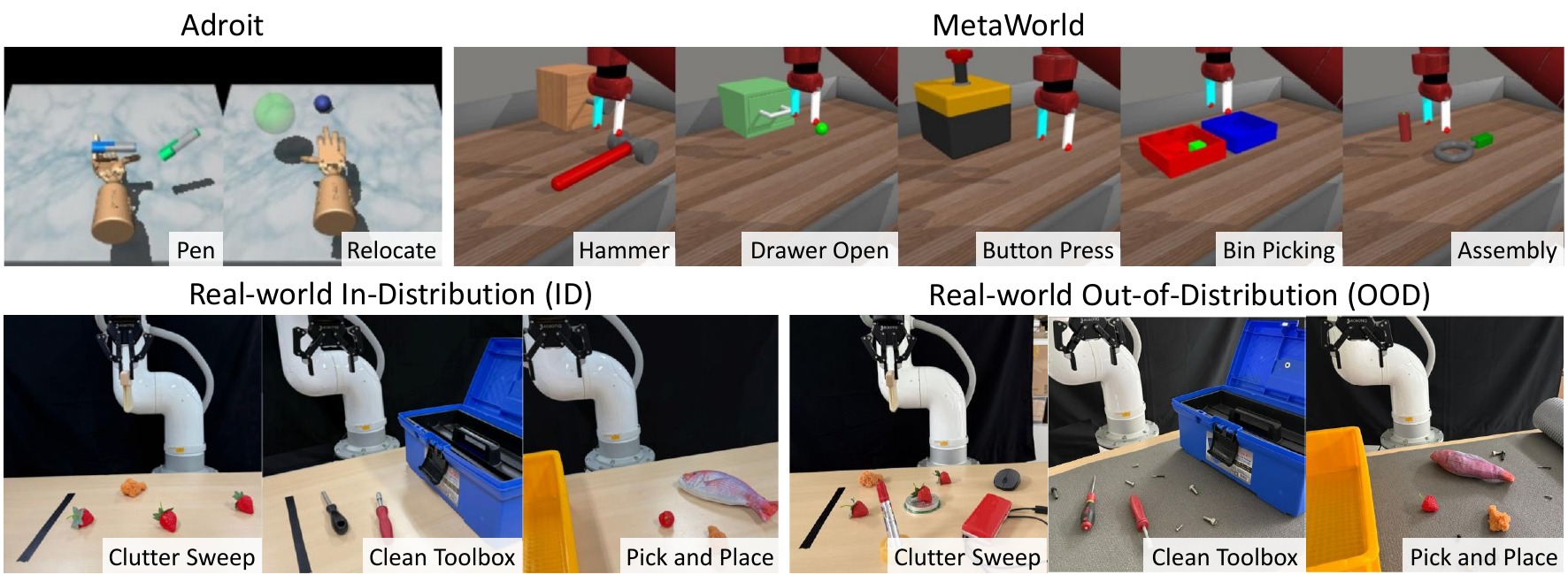}
\end{center}
\vspace{-0.5cm}
\caption{
\textbf{Visual illustration of the simulation and real-world benchmarks.}
We adopt the Adroit~\cite{rajeswaran2018adroit} and MetaWorld~\cite{yu2020metaworld} tasks as our simulated environments, and clutter sweep, clean toolbox, and pick-and-place as our real-world tasks.
We further evaluate under visual distribution shifts, including desk mat and lighting changes and increased clutter.
}
\vspace{-0.4cm}
\label{fig:env_vis}
\end{figure*}

\vspace{-0.3cm}
\paragraph{Student Architecture.}
The student network uses a single RGB encoder 
$f^{\mathrm{S}}_{\mathrm{RGB}}$
and a policy head that shares the teacher's architecture but is randomly initialized.
Crucially, the student operates using \textbf{only} the raw RGB image $o_t$ and proprioception $p_t$. By eliminating the dependency on SAM~2 and DAM~2 during inference, the student drastically reduces latency and removes the source of potential visual instability.

\vspace{-0.3cm}
\paragraph{Student Objective.}
To enable the student to make robust decisions from RGB alone, we explicitly train it to replicate the teacher's internal representations. We employ a feature alignment loss that maximizes the cosine similarity between 
the student's embedding $z_t$ and each of the teacher's modality-specific embeddings ($z_t^{\mathrm{RGB}}, z_t^{\mathrm{object}}, z_t^{\mathrm{depth}}$):
\begin{equation}
\mathcal{L}_{\text{feature}} = \sum_{m \in \{\text{RGB, object, depth}\}} \left( 1 - \frac{\text{sg}[z_t^m] \cdot z_t}{\| \text{sg}[z_t^m] \| \| z_t \|} \right).
\label{eq: feature loss}
\end{equation} 
where $\mathrm{sg}[\cdot]$ denotes the stop-gradient operator.
Minimizing this loss encourages the student's RGB encoder to effectively capture the geometric and semantic features that the teacher derived from the privileged VFM inputs.

The final training objective for the student is 
a weighted sum of the behavior cloning loss on expert actions, and the feature distillation loss:
\begin{equation}
\mathcal{L}_\text{student} = \mathcal{L}_\text{BC} + \lambda \mathcal{L}_\text{feature},
\label{eq: student loss}
\end{equation}
where $\lambda$ balances behavior cloning with feature alignment.
We anneal $\lambda$ from 1 to 0 (Supp.~A.5), so late training reduces to standard behavior cloning. A fixed $\lambda$ performs comparably (\Cref{tab:distill_config}), indicating the gain comes from the teacher cues rather than the schedule.
By optimizing this objective, the student learns a robust, low-latency policy that retains the superior performance characteristics of the VFM-augmented teacher.

\section{Experiments}
\label{section: experiment_setup}

In the following sections, we aim to answer the following research questions:
\begin{itemize}
    \item Can AFO transfer the privileged visual representations of VFMs to a lightweight policy? (\Cref{subsection: main experiment})
    \item Does AFO preserve the VFMs' robustness under visual distribution shifts? (\Cref{subsection: out_of_distribution_generalization})
    \item Why is directly deploying the VFM-based teacher problematic for closed-loop control? (\Cref{subsection: qualitative_analysis})
    \item How do depth and segmentation cues each contribute to AFO? (\Cref{subsection: ablation studies})
\end{itemize}

\subsection{Environmental Setup}

\paragraph{Simulation Benchmarks.}

We evaluate AFO on seven tasks drawn from two widely used manipulation benchmarks in simulation (\Cref{fig:env_vis}(a,b)):
\begin{itemize}
    \item \textbf{Adroit}~\cite{rajeswaran2018adroit}: A high-dimensional dexterous manipulation domain using a 24-DoF shadow hand. We select two difficult tasks: \textit{Reorient-Pen} and \textit{Relocate}, which require precise contact management to manipulate objects to target poses.
    \item \textbf{MetaWorld}~\cite{yu2020metaworld}: A tabletop manipulation suite with a Sawyer arm. Following prior work~\cite{nair2022r3m,majumdar2023vc1}, we select five distinct tasks, \textit{Hammer, Assembly, Bin Picking, Button Press, Drawer Open}, that evaluate reaching, grasping, and tool-use capabilities.
\end{itemize}

\vspace{-0.3cm}
\paragraph{Real-world Tasks.}

In addition, for real-world experiments, we evaluate with a 6-DoF Indy7 collaborative arm equipped with a Robotiq 2F85 gripper.
The setup includes two Intel RealSense D435 cameras (wrist-mounted and third-person). 
We design three tasks requiring precise geometry understanding (\Cref{fig:env_vis}(c)):
\begin{itemize}
    \item \textbf{Clutter Sweeping:} The robot must distinguish target strawberries from distractors and sweep them past a designated line.
    \item \textbf{Clean Toolbox:} A long-horizon task involving opening a toolbox, removing a divider, placing two screwdrivers inside, and closing the lid.
    \item \textbf{Pick-and-Place:} The robot picks three specific objects and places them into a bin.
\end{itemize}

\vspace{-0.3cm}
\paragraph{Baselines.}

We compare AFO against a diverse set of visual representation learning baselines, including scratch-trained models, large-scale pre-trained encoders, and recent foundation-model distillation approaches.

\begin{itemize}
    \item \textbf{LfS}~\cite{hansen2022lfs}: A strong baseline trained from scratch with data augmentation~\cite{laskin2020rad}.
    \item \textbf{MVP}~\cite{radosavovic2022mvp}: A Vision Transformer pretrained via masked autoencoding on large-scale egocentric video data.
    \item \textbf{VC-1}~\cite{majumdar2023vc1}: A ViT-based encoder pretrained on a massive video corpus using MAE.
    \item \textbf{R3M}~\cite{nair2022r3m}: A representation learned through temporal consistency objectives on human interaction videos.
    \item \textbf{DINOv2}~\cite{oquab2023dinov2}: A self-supervised foundation model producing object-centric features.
    \item \textbf{Theia}~\cite{shang2024theia}: A foundation-model distillation approach that fuses CLIP~\cite{radford2021clip}, SAM~\cite{kirillov2023sam}, and DAM~\cite{depth_anything_v1} into a single backbone. Theia performs distillation on ImageNet~\cite{ImageNet}, introducing a domain gap for robot control~\cite{mehraunderstanding2024gap}.
\end{itemize}
\vspace{-0.2cm}

\vspace{-0.2cm}
\paragraph{Implementation Details.}

For simulation experiments, each observation consists of a $256 \times 256$ RGB image and proprioceptive states, which are concatenated with visual features before being passed to MLP policy head~\cite{hansen2022lfs, majumdar2023vc1}. Following prior work~\cite{nair2022r3m, majumdar2023vc1, parisi2022unsurprising, hansen2022lfs, hwang2024coin}, we train policies using 100 expert demonstrations per task in Adroit and 25 demonstrations per task in MetaWorld, and report results averaged over five random seeds.

For real-world experiments, all policies are trained only on downstream robotic demonstrations, with 35 demonstrations per task. We use an action chunking transformer policy~\cite{zhao2023ACT}, where methods share the same transformer policy head but differ in the visual encoder. The policy predicts task-space end-effector commands and gripper actions from visual observations and proprioceptive inputs, and the robot executes actions at 20Hz. Each policy is evaluated over 10 rollouts per seed under identical initial conditions across methods. Additional training and evaluation details are provided in the supplementary material.

\begin{table*}[ht]
\caption{
\textbf{AFO outperforms all baselines at minimal computational cost.}
AFO's student matches or surpasses VFM-based baselines on Adroit and MetaWorld while running at the lowest latency and parameter count, on par with the smallest from-scratch baseline (LfS).
}
\vspace{-0.65cm}
\label{table: main result}
\begin{center}
\resizebox{\textwidth}{!}{%
\begin{tabular}{l|M{1.2cm}M{1.2cm}|M{1.2cm}M{1.2cm}M{1.2cm}M{1.2cm}M{1.2cm}|M{1.8cm}|cc}
\toprule
\multirow{3}{*}[0.1cm]{Method} & \multicolumn{2}{c|}{Adroit~\cite{rajeswaran2018adroit}} & \multicolumn{5}{c|}{MetaWorld~\cite{yu2020metaworld}} & \multirow{3}{*}[0.1cm]{\begin{tabular}{c}Average \\ (7 tasks)\end{tabular}} &
\multicolumn{2}{c}{Efficiency} \\
\cmidrule(lr){2-3}\cmidrule(lr){4-8}\cmidrule(lr){10-11} \\[-2.5ex]
& Pen & Relocate & Assembly & Bin & Button & Drawer & Hammer & & Latency (ms) & \# Params \\
\drule
\multicolumn{11}{l}{\textit{Use Pretrained VFMs at inference time.}}\\[0.3ex]
DINOv2~\cite{oquab2023dinov2} & 72.80 & 59.20 & \textbf{100.00} & 89.60 & 89.60 & \textbf{100.00} & 98.40 & 87.09\std{1.60} & 12.48 & 88M \\
R3M~\cite{nair2022r3m} & 72.00 & 70.40 & \textbf{100.00} & 88.80 & \textbf{96.80} & \textbf{100.00} & 96.00 & \underline{89.14}\std{0.57} & \underline{6.41} & \underline{26M} \\
MVP~\cite{radosavovic2022mvp} & 68.00 & 56.80 & \textbf{100.00} & 95.20 & 75.20 & \textbf{100.00} & 98.40 & 84.80\std{0.96} & 7.11 & 87M \\
VC-1~\cite{majumdar2023vc1} & 68.80 & 52.80 & \textbf{100.00} & 93.60 & 75.20 & \textbf{100.00} & \textbf{100.00} & 84.34\std{0.87} & 6.84 & 87M \\
Theia~\cite{shang2024theia} & 72.00 & 65.60 & \textbf{100.00} & 91.20 & 84.00 & \textbf{100.00} & \underline{99.20} & 87.43\std{2.74} & 7.94 & 88M \\
\model~(Teacher) & 69.60 & \underline{75.20} & 75.20 & \textbf{98.40} & 92.00 & 99.20 & \textbf{100.00} & 87.09\std{1.54} & 110.07 & 8.2B \\
\textcolor{gray!85}{\model~(Teacher w/ GT)} & \textcolor{gray!85}{76.00} & \textcolor{gray!85}{96.80} & \textcolor{gray!85}{94.40} & \textcolor{gray!85}{94.40} & \textcolor{gray!85}{99.20} & \textcolor{gray!85}{100.00} & \textcolor{gray!85}{100.00} & \textcolor{gray!85}{94.40\std{1.30}} & \textcolor{gray!85}{3.84} & \textcolor{gray!85}{3.0M} \\

\midrule

\multicolumn{11}{l}{\textit{No Pretrained VFMs at inference time.}}\\[0.3ex]
LfS~\cite{hansen2022lfs} & \underline{74.40} & 48.80 & 76.00 & \underline{96.80} & \underline{96.00} & \textbf{100.00} & \textbf{100.00} & 84.57\std{2.02} & \textbf{2.11} & \textbf{2.7M} \\
\model~(Student) & \textbf{79.20} & \textbf{84.00} & \underline{92.00} & \underline{96.80} & \underline{96.00} & \textbf{100.00} & \textbf{100.00} & \textbf{92.57}\std{1.51} & \textbf{2.11} & \textbf{2.7M} \\

\bottomrule

\end{tabular}}
\end{center}
\vspace{-0.5cm}
\end{table*}

\begin{table*}[t]
\centering
\caption{
\textbf{Comparison on Real-world Imitation Learning.}
Success rates under in-distribution (ID) and visual out-of-distribution (OOD) conditions on three real-world tasks.
While all methods perform comparably in distribution, AFO degrades far less under visual shifts, surpassing both baselines by a large margin in OOD at the same cost as the lightweight ACT.
}
\label{table: real_world_split}
\vspace{-0.3cm}
\resizebox{0.80\textwidth}{!}{%
\begin{tabular}{M{1.5cm}|l|ccc|c|cc}
\toprule
\multirow{2}{*}[-0.6ex]{\shortstack[c]{Problem\\[0.2ex]Setting}} & \multirow{2}{*}[-0.6ex]{Method} & \multicolumn{3}{c|}{Real-world} & \multirow{2}{*}[-0.1cm]{\begin{tabular}{c}Average \\ (3 tasks)\end{tabular}} & \multicolumn{2}{c}{Efficiency} \\
\cmidrule(lr){3-5}\cmidrule(lr){7-8}
 &  & Clutter Sweep & Clean Toolbox & Pick \& Place &   & Latency (ms) & \# Params \\
\drule
\multirow{3}{*}{ID}
& ACT~\cite{zhao2023ACT}   & \underline{68.89} & 47.50 & \underline{60.00} & \underline{58.80}\std{8.45} & \textbf{4.32} & \textbf{34.19M} \\
& Theia~\cite{shang2024theia} & 66.70 & \textbf{50.00} & 58.90 & 58.53\std{4.88} & 7.94 & 125.24M \\
& AFO   & \textbf{74.40} & \underline{49.17} & \textbf{70.00} & \textbf{64.52}\std{2.16} & \textbf{4.32} & \textbf{34.19M} \\
\midrule
\multirow{3}{*}{OOD}
& ACT~\cite{zhao2023ACT}   & \underline{48.88} & \underline{43.33} & 37.78 & \underline{43.33}\std{5.14} & \textbf{4.32} & \textbf{34.19M} \\
& Theia~\cite{shang2024theia} & 44.44 & 41.66 & \underline{43.33} & 43.14\std{3.36} & 7.94 & 125.24M \\
& AFO   & \textbf{60.00} & \textbf{46.67} & \textbf{51.11} & \textbf{52.59}\std{5.30} & \textbf{4.32} & \textbf{34.19M} \\
\bottomrule
\end{tabular}
}
\vspace{-0.5cm}
\end{table*}

\vspace{0.1cm}
\subsection{Main Results}
\label{subsection: main experiment}
\paragraph{Distillation improves performance without increasing computational cost.}

In \Cref{table: main result}, we first evaluate whether AFO can transfer privileged VFM knowledge into a lightweight RGB-only policy (\ie, AFO Student).
Compared with LfS~\cite{hansen2022lfs}, which trains from scratch without distillation, AFO Student substantially improves the success rate from 48.80\% to 84.00\% on Adroit Relocate and from 84.57\% to 92.57\% on average.
Since AFO Student shares the same architecture and inference cost as LfS, this gain is attributable to distillation alone, and indicates that depth and segmentation cues from VFMs provide useful semantic and geometric supervision beyond what can be learned from task demonstrations alone.

\vspace{-0.3cm}
\paragraph{AFO Student outperforms VFM-based policies without using VFMs.}
As shown in \Cref{table: main result}, AFO Student achieves 92.57\% on the 7-task average, outperforming the strongest VFM-based baselines, R3M~\cite{nair2022r3m} and Theia~\cite{shang2024theia}, by 3.43 and 5.14 while running without any VFM at inference.
R3M benefits from robot-relevant visual pretraining but does not use dense VFM cues such as segmentation masks and depth maps.
Theia is the most directly related baseline, as it also distills from multiple foundation models, but it distills on generic image data (\ie, ImageNet~\cite{ImageNet}), whereas AFO distills VFM-derived cues during robotic policy learning, grounding the learned features in manipulation.

Notably, AFO Student also surpasses its own AFO Teacher, despite using only RGB at inference.
This does not mean that VFM cues are unhelpful; they provide useful supervision during training, but deploying the teacher directly remains sensitive to VFM prediction errors and temporal inconsistencies, which we analyze in \Cref{subsection: qualitative_analysis}.

\vspace{-0.3cm}
\paragraph{Computational Cost Comparison.}

AFO improves performance while preserving the efficiency required for closed-loop control. As shown in \Cref{table: main result}, AFO Student uses only 2.7M parameters and runs at 2.11ms, matching the inference cost of LfS. In contrast, pretrained-backbone methods require substantially larger models: R3M uses 26M parameters, while DINOv2, MVP, VC-1, and Theia use ViT-B-scale backbones with around 88M parameters. These larger encoders also incur higher latency, making them less suitable for high-frequency visuomotor control.

The cost becomes even more prohibitive when VFMs are directly deployed as part of the teacher policy. AFO Teacher requires running multiple VFMs for depth and segmentation, resulting in 8.2B parameters and 110.07ms latency. Such latency can slow down the perception–action loop and make closed-loop execution less responsive, especially in contact-rich or long-horizon manipulation tasks. By distilling VFM-derived cues into a compact RGB-only student, AFO avoids this inference-time overhead while retaining the useful supervision provided by VFMs during training.

\begin{figure*}[t]
\begin{center}
\includegraphics[width=\textwidth]{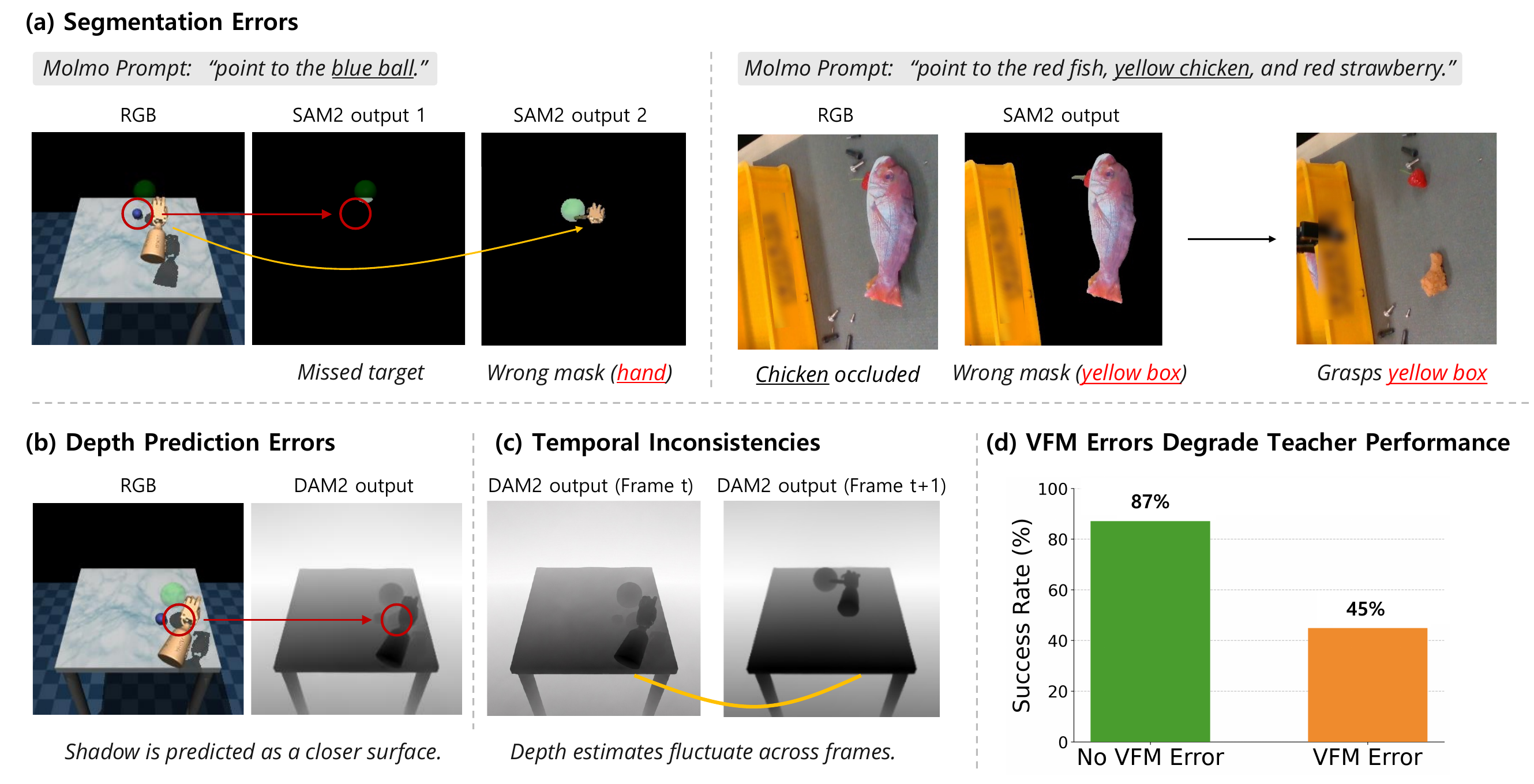}
\end{center}
\vspace{-0.65cm}
\caption{
\textbf{Failure Analysis of VFM-Based Teacher Policies.}
Directly using VFMs in the teacher introduces brittle failure modes.
\textbf{(a)} Segmentation errors: SAM~2 misses the target or masks an irrelevant object.
\textbf{(b)} Depth errors: DAM~2 mis-estimates geometry, \eg, predicting a shadow as a closer surface.
\textbf{(c)} Temporal inconsistencies: depth estimates fluctuate across adjacent frames.
\textbf{(d)} These errors substantially degrade performance: on Adroit Relocate, the teacher succeeds 87\% without VFM errors, but only 45\% when an error occurs.
}
\label{fig:fig_why}
\vspace{-0.45cm}
\end{figure*}

\subsection{Out-of-Distribution Generalization}
\label{subsection: out_of_distribution_generalization}

We next evaluate whether the distilled student inherits visual generalization from VFM-derived supervision in real-world manipulation.
As shown in \Cref{table: real_world_split}, 
under 
in-distribution (ID) conditions, where the workspace, lighting, and object configurations stay close to the demonstrations,
AFO achieves the highest average of 64.52\%, outperforming ACT by 5.72 points and Theia by 5.99 points, showing that VFM-derived depth and segmentation cues provide useful supervision for real-world imitation learning.
We then
evaluate the policies under out-of-distribution (OOD) conditions with visual and geometric shifts, including changes in desk mats and lighting, increased clutter and distractors, and more randomized object placements, as illustrated in \Cref{fig:env_vis}.  Under these conditions, AFO achieves the highest success rate on all three tasks: 60.00\% on Clutter Sweep, 46.67\% on Clean Toolbox, and 51.11\% on Pick \& Place. Its average success rate improves over ACT from 43.33\% to 52.59\% and over Theia from 43.14\% to 52.59\%. 

These results indicate that AFO improves real-world manipulation not only under familiar conditions, but also under visual distribution shifts. In the OOD setting, changes in background, lighting, clutter, and object placement can make policies more sensitive to irrelevant visual variations. The depth and segmentation cues provided by VFMs are useful in this setting because they emphasize task-relevant geometry and objects. AFO transfers these cues to the RGB-only student during training, enabling the deployed policy to benefit from VFM-level visual supervision without directly running VFMs during closed-loop execution.

\subsection{Failure Modes of VFM Teachers}
\label{subsection: qualitative_analysis}

Previous section shows that AFO outperforms all baselines.
However, it may appear counterintuitive that the distilled student surpasses its own teacher.
This section analyzes how the failure modes of the VFMs correlate with this behavior, explaining why directly deploying the teacher is both error-prone and computationally expensive, and why distillation alleviates these issues.
As a first indication, \Cref{table: main result} shows that AFO (Teacher w/ GT), a teacher 
given privileged ground-truth segmentation and depth in simulation, outperforms every baseline, suggesting that teacher errors, rather than the modalities themselves, limit its performance.

\vspace{-0.3cm}
\paragraph{VFM predictions are unreliable on downstream robotic data.}
Because VFMs are general-purpose foundation models, their predictions are not always reliable on downstream robotic observations, leading to occasional errors that propagate into the teacher policy.
In simulation, we observe two recurring error types (\Cref{fig:fig_why} (a, b)): \textit{segmentation errors}, where SAM~2 fails to segment the target object or instead segments irrelevant regions such as the robot hand or table; and \textit{depth errors}, where DAM~2 misestimates object distances, for instance by interpreting shadows as nearby surfaces.
To quantify their impact, we analyze the teacher on the Adroit \textit{Relocate} task.
% , which exhibits the lowest teacher success rate.

Across 150 rollouts, we manually label each trajectory by whether both SAM~2 and DAM~2 remain accurate throughout the episode or at least one misprediction occurs (\Cref{fig:fig_why} (d)).
When no VFM error occurs, the teacher succeeds in 87\% of rollouts (88/101); a single misprediction drops this to 45\% (22/49). 
This correlational evidence, together with the controlled GT-teacher comparison in Table~\ref{table: main result}, reveals the teacher's acute sensitivity to VFM reliability.
This brittleness also appears in the real world. In Pick-and-Place, Molmo is run only at the initial timestep for efficiency (\Cref{subsection: teacher training}). When the target object is heavily occluded at $t{=}0$, Molmo misidentifies a yellow box as the target ``yellow chicken,'' causing SAM~2 to segment the wrong object. Despite accurate RGB and depth, this single error derails the entire trajectory.

\vspace{-0.3cm}
\paragraph{Static per-frame inference induces temporal inconsistency.}
Since DAM 2 performs inference independently on each frame and SAM 2 propagates its mask by video tracking which may fail under occlusion, their predictions can be temporally unstable across a video stream.
For static scene elements such as table corners, depth estimates fluctuate across frames despite unchanged scene geometry (\Cref{fig:fig_why} (c)).
Such jitter, together with the high inference latency of SAM~2 and DAM~2, introduces jerky motions that further degrade performance in real-world tasks such as Clean Toolbox.

\vspace{-0.3cm}
\paragraph{Distillation mitigates these failures.}
We attribute the student's advantage to distillation combined with end-to-end training on downstream data.
Rather than consuming raw VFM outputs at every step, the student learns from the teacher's behavior and internal features while adapting its RGB representation end-to-end to the target domain.
This lets it retain the semantic benefits of VFMs while avoiding both their per-frame errors and temporal instability, as well as their computational overhead, yielding more robust and reliable behavior at deployment, and, we conjecture, the higher final performance observed in \Cref{subsection: main experiment}.

\vspace{0.15cm}
\subsection{Ablation Studies}
\label{subsection: ablation studies}

\paragraph{Contribution of Depth and Segmentation Cues.}
We examine how depth and segmentation contribute to AFO (\Cref{table: All_stage}). Adding either cue improves over the RGB-only teacher, showing that both provide useful information. However, adding segmentation to depth does not further improve the teacher. We observed frequent segmentation failures in tasks such as MetaWorld Assembly, with multiple small task-relevant objects, making the teacher sensitive to erroneous SAM outputs. In contrast, the student performs best when distilled from both cues, reaching 92.57\% compared with 89.26\% and 91.54\% from depth-only and segmentation-only teachers. This indicates that both cues provide complementary supervision through distillation, while the student avoids direct reliance on erroneous VFM outputs at inference. The strong Teacher w/ GT performance  shows that both cues are beneficial when accurate.

\vspace{-0.3cm}
\paragraph{Distilling VFM Outputs Beats Distilling VFM Features.}
Theia~\cite{shang2024theia} also distills from multiple foundation models, but it distills \textit{intermediate VFM features} on \textit{generic image data} (\ie, ImageNet~\cite{ImageNet}), neither of which is optimized for manipulation: intermediate features retain background and distractor information, and generic data provides no task grounding.
AFO instead distills VFM \textit{outputs}, namely segmentation masks and depth maps, on downstream demonstrations.
\Cref{tab:comparison_theia} isolates the two factors: moving to downstream data improves the 7-task average from 87.43\% to 90.86\%, and switching the target to VFM outputs adds a further 5.71 points.
Our CNN student reaches 92.57\% with only 2.72M parameters, exceeding both ViT-based feature-distillation variants.

\begin{table}[t]
  \centering
  \caption{
  \textbf{Effect of teacher modalities on distillation.}
  Depth and segmentation each improve the teacher, but teacher accuracy does not
  predict distillation quality: only the teacher with both modalities yields the
  best RGB-only student (2.11 ms, 2.72M).
  }
  \vspace{-0.2cm}
  \resizebox{\columnwidth}{!}{%
  \begin{tabular}{ccc|cc|cc}
  \toprule
  \multicolumn{3}{c|}{Teacher Modality} & \multicolumn{2}{c|}{Teacher Cost} & \multicolumn{2}{c}{Average (7 tasks)} \\
  \cmidrule(lr){1-3} \cmidrule(lr){4-5} \cmidrule(lr){6-7}
  RGB & Depth & Segment. & Latency (ms) & \# Params. & Teacher & Student \\
  \drule
  \cmark & --     & --     & 2.11   & 2.72M  & 84.57\std{2.02} & -- \\
  \cmark & \cmark & --     & 42.78  & 100.3M & \textbf{88.34}\std{1.65} & 89.26\std{3.49}  \\
  \cmark & --     & \cmark & 70.95  & 8.10B  & 85.37\std{4.51} & \ul{91.54}\std{1.48} \\
  \cmark & \cmark & \cmark & 110.07 & 8.20B  & \ul{87.09}\std{1.54} & \textbf{92.57}\std{1.51} \\
  \bottomrule
  \end{tabular}
  }
  \label{table: All_stage}
  \vspace{-0.3cm}
\end{table}

\begin{table}[t]
\centering
\caption{\textbf{Comprehensive Comparison with Theia Distillation Objective.} The first row is the original Theia and the last row is AFO (Student); the two middle rows vary one factor at a time.}
\label{tab:comparison_theia}
\vspace{-0.2cm}
\resizebox{\linewidth}{!}{%
\begin{tabular}{ccc|cc|c}
\toprule
\multicolumn{3}{c|}{Distillation Configuration} &
\multirow{2}{*}[-0.1cm]{\begin{tabular}{c}Adroit \\ (2 tasks)\end{tabular}}  &
\multirow{2}{*}[-0.1cm]{\begin{tabular}{c}MetaWorld \\ (5 tasks)\end{tabular}}  &
\multirow{2}{*}[-0.1cm]{\begin{tabular}{c}Average \\ (7 tasks)\end{tabular}}
\\
\cmidrule(lr){1-3}
Dataset & Distill Target & Capacity & 
&  & \\
\drule

\cellcolor{gray!20}\textit{ImageNet} & \cellcolor{gray!20}VFM \textit{Features} & \cellcolor{gray!20}\textit{ViT} &
{68.80}\std{5.40} & {94.88}\std{2.37} & 87.43\std{2.74} \\

Downstream & \cellcolor{gray!20}VFM \textit{Features} & \cellcolor{gray!20}\textit{ViT} &
{80.80}\std{5.02} &  {94.88}\std{2.09} & 90.86\std{1.40} \\

Downstream & VFM \textit{Outputs} & \cellcolor{gray!20}\textit{ViT} &
\textbf{88.80}\std{3.03} & \textbf{99.68}\std{0.44} & \textbf{96.57}\std{0.81} \\

Downstream & VFM \textit{Outputs} & CNN &
\ul{81.60}\std{1.67} & \ul{96.96}\std{1.54} & \ul{92.57}\std{1.51} \\

\bottomrule

\end{tabular}}
\vspace{-0.4cm}
\end{table}

\vspace{-0.3cm}
\paragraph{Distillation Configurations.}
\Cref{tab:distill_config} extends this analysis to the student.
Distilling from a teacher without segmentation or without depth degrades performance in both cases, confirming that the two cues contribute complementary information that survives distillation.
In contrast, the loss function, alignment strategy, and $\lambda$ schedule all remain close to our default, indicating that the gains come from the teacher cues rather than the specific distillation objective.

\begin{table}[t]
\centering
\caption{\textbf{Comprehensive Ablation Studies.} Removing a teacher modality costs more than changing the distillation configuration: dropping segmentation degrades performance noticeably, whereas the loss, alignment, and $\lambda$ variants all remain close to our default.}
\label{tab:distill_config}
\vspace{-0.3cm}
\resizebox{\linewidth}{!}{%
\begin{tabular}{M{1.3cm}M{1.3cm}|M{1.3cm}M{2.0cm}M{1.3cm}|M{1.5cm}}
\toprule
\multicolumn{2}{c|}{Teacher} & \multicolumn{3}{c|}{Distillation Configuration} & \multirow{2}{*}[-0.1cm]{\begin{tabular}{c}Average \\ (7 tasks)\end{tabular}}  \\
\cmidrule(lr){1-2}\cmidrule(lr){3-5}
Segment. & Depth & Loss & Alignment & $\lambda$ & \\
\drule
\multicolumn{5}{c|}{\emph{BC-only from scratch (LfS), no distillation}} & 84.57\std{2.02} \\
\midrule
\cellcolor{gray!20}\emph{--} & DAM & cos & per-modality & dynamic & 89.26\std{3.49} \\
SAM & \cellcolor{gray!20}\emph{--} & cos & per-modality & dynamic & 91.54\std{1.48} \\
\midrule
SAM & DAM & \cellcolor{gray!20}\emph{$\ell_2$} & per-modality & dynamic & \ul{92.00}\std{2.49} \\
SAM & DAM & cos & \cellcolor{gray!20}\emph{fused} & dynamic & \ul{92.00}\std{1.46} \\
SAM & DAM & cos & per-modality & \cellcolor{gray!20}\emph{fixed} 1.0 & 91.43\std{2.10} \\
\midrule
SAM & DAM & cos & per-modality & dynamic & \textbf{92.57}\std{1.51} \\
\bottomrule
\end{tabular}}
\vspace{-0.4cm}
\end{table}

\section{Conclusion}
\label{section: conclusion}

We presented {\model}, a two-stage privileged learning framework that uses vision foundation models (VFMs) as privileged information for visuomotor control.
We fuse depth from DAM~2~\cite{depth_anything_v2} and masks from SAM~2~\cite{ravi2024sam2}, localized via Molmo's~\cite{deitke2025molmo} pointing, to train a teacher, then distill it into a lightweight, RGB-only student that needs no VFMs at inference.
Across seven simulated and three real-world tasks, the student surpasses its teacher, stays robust under visual distribution shifts, and runs at a fraction of its latency and memory cost.
VFMs thus serve better as privileged training signals than as in-loop components.

\vspace{-0.3cm}
\paragraph{Limitations and Future Work.}
Our study focuses on policies with lightweight visual encoders rather than large backbones such as VLAs.
While this favors efficiency and real-time deployability, it may limit the absolute performance and generalization that larger backbones offer.
Extending privileged VFM distillation to VLAs~\cite{black2024pi0, black2025pi05, liu2024rdt, geminirobotics2025} is thus a promising direction: a VLA could be trained with segmentation and depth cues from SAM and DAM, which are then distilled away to retain the gains without inference-time overhead.
Since teacher usefulness varies across tasks, adapting the distillation weights to task relevance could further improve performance.

\section*{Acknowledgements}
This work was supported by the Industrial Technology Innovation Program (P0028404) of the Ministry of Trade, Industry and Resource of Korea; the Institute for Information \& communications Technology Planning \& Evaluation (IITP) grant funded by the Korea government (MSIT) (RS-2019-II190075, Artificial Intelligence Graduate School Program (KAIST)); the National Research Foundation of Korea (NRF) grant funded by the Korea government (MSIT) (No. RS-2025-00555621); and the ``Advanced GPU Utilization Support Program'' funded by the Government of the Republic of Korea (Ministry of Science and ICT).

{
    \small
    \bibliographystyle{ieeenat_fullname}
    \bibliography{IEEEabrv,reference}
}

%% file: supplementary.tex
\title{All-For-One: Distilling Visual Foundation Models \\ to a Fast, Memory-Efficient Visuomotor Policy\\
--- Supplementary Material ---}

\maketitle

\setcounter{section}{0}
\renewcommand{\thesection}{\Alph{section}}
\renewcommand{\thetable}{\Alph{section}\arabic{table}}
\renewcommand{\thefigure}{\Alph{section}\arabic{figure}}
\setcounter{table}{0}
\setcounter{figure}{0}

In this supplementary material, we provide additional implementation details,
evaluation protocols, quantitative analyses, and qualitative results to
complement the main paper.
In Sections~\ref{supp:sim_training}--\ref{supp:real_training}, we describe the
simulation and real-world training setups, including model architectures,
trainable and frozen components, and optimization details.
Sections~\ref{supp:real_eval}--\ref{supp:real_ood} detail the real-world evaluation
protocol, task scoring criteria, and visual distribution shifts.
Sections~\ref{appendix: model architecture}--\ref{appendix: molmo prompts} provide
additional details on the distillation procedure, hyperparameters, and Molmo
prompts used for SAM~2.
We then describe the latency measurement protocol, the annotation criteria for
Fig.~4(d), and the statistical analysis in
Sections~\ref{supp:latency}--\ref{supp:significance}.
Finally, Section~\ref{supp:full_results} reports complete per-task results and
ablation studies, while Section~\ref{supp:qualitative} provides qualitative VFM
outputs across the Adroit and MetaWorld tasks.

\startcontents[appendices]
\printcontents[appendices]{l}{1}{\setcounter{tocdepth}{2}}

\section{Additional Implementation Details \\ and Experimental Settings}
\label{appendix: implementation_details}

\subsection{Simulation Training}
\label{supp:sim_training}

\paragraph{Architecture and Inputs/Outputs}
Each input modality (RGB, Depth, Segment) is encoded by a separate lightweight convolutional encoder following~\cite{hansen2022lfs}. The resulting feature maps are summed before the policy head.
For ViT baselines~\cite{oquab2023dinov2, radosavovic2022mvp, majumdar2023vc1, shang2024theia}, we use spatial tokens (not the \texttt{[CLS]} token) as policy input, following~\cite{yadav2023ovrl,shang2024theia}.
For all simulation experiments, observations are resized to $256 \times 256$ RGB images.
Proprioceptive states, such as joint angles, are concatenated with the visual feature before being passed to the policy head.
All methods use the same policy structure consisting of a visual encoder and a lightweight policy head. For pretrained baselines, the visual backbones are frozen while the downstream lightweight encoders and policy heads are trained. In AFO, SAM 2~\cite{ravi2024sam2} and DAM 2~\cite{depth_anything_v2} are likewise frozen, while the modality encoders and policy head are trained.

\paragraph{Evaluation Protocol}
We train each policy for 100 epochs using 100 expert demonstrations per task in Adroit~\cite{rajeswaran2018adroit} and 25 demonstrations per task in MetaWorld~\cite{yu2020metaworld}.
Policies are evaluated every 5 epochs from epoch 50 to epoch 100, and the best-performing checkpoint is selected for each run following~\cite{majumdar2023vc1}.
Each evaluation runs 25 rollouts per task, and final results are averaged over five random seeds.

\subsection{Real-World Training}
\label{supp:real_training}

For real-world experiments, all policies are trained  on downstream robotic demonstrations. We collect 35 demonstrations per task and train policies for 1000 epochs, except for Clean Toolbox, which is trained for 1500 epochs due to its longer horizon. 
All methods use an action chunking transformer policy. For ACT~\cite{zhao2023ACT} and AFO, the 5-layer CNN encoders used in simulation are replaced by trainable ResNet-18 encoders~\cite{he2016resnet}.
AFO Teacher uses three separate trainable ResNet-18 encoders for RGB, depth, and segmentation, while SAM 2 and Depth Anything V2 remain frozen.
AFO Student uses a single trainable ResNet-18 RGB encoder.
For Theia~\cite{shang2024theia}, the pretrained ViT-B~\cite{dosovitskiy2020image} is frozen in simulation but fine-tuned during real-world policy training, following the fine-tuned real-robot setting in the original paper~\cite{shang2024theia}.
All policy heads are trained, and random shift augmentation is applied to the visual inputs during training. Each policy is trained across three random seeds.

\subsection{Real-World Evaluation}
\label{supp:real_eval}

The policy predicts task-space actions in a relative end-effector trajectory representation~\cite{chi2024umi}, \ie, delta end-effector position and orientation together with the gripper state, conditioned on visual observations and proprioceptive inputs consisting of joint positions and velocities.
The robot executes commands at 20\,Hz using action chunks of length 20, where 10 actions are executed from each chunk. 
Each trained policy is evaluated with 10 rollouts per seed using identical initial conditions across methods, giving 30 rollouts per method and task across the three seeds.

Success and failure criteria are defined per task, applied identically to all methods, and fixed
before evaluation. Each rollout is scored by task progress rather than binary success. Clutter
Sweep and Pick-and-Place are scored out of three stages and Clean Toolbox out of four, so a
rollout receives a score in $\{0, 1/3, 2/3, 1\}$ and $\{0, 1/4, 2/4, 3/4, 1\}$ respectively, and
reported success rates are the mean of these per-rollout scores. For Clutter Sweep, one point is
given for each target strawberry swept across the black line, and the episode is terminated as a
failure if non-target objects are swept across the line, if the episode exceeds 30 seconds, or if
the robot sweeps empty space more than five times. For Pick-and-Place, one point is given for
each of the three target objects placed in the bin in any order, and the episode is terminated as
a failure if the robot fails to grasp a target object in three consecutive attempts. For Clean
Toolbox, one point is given for opening the lid, one for each of the two screwdrivers placed
inside, and one for closing the lid, and the episode is terminated if the policy fails any stage
five times. For these multi-stage tasks, the success rate reported in the main paper is defined as the mean normalized task-completion score.

\subsection{Real-World Visual Shifts}
\label{supp:real_ood}
For in-distribution evaluation, we use the same task setup as the demonstrations while varying initial object configurations within the demonstrated range.
For out-of-distribution evaluation, we introduce additional visual and geometric shifts, including changes in desk mats and lighting, increased clutter and distractors, and more randomized object placements. For each task and seed, the 10 rollouts individually vary desk mats (2 rollouts), lighting (2), clutter and distractors (2), and object placements (2), with all shifts combined in the remaining 2. All policies are evaluated under these conditions without additional training or fine-tuning. These in-distribution and out-of-distribution conditions are illustrated in Figure~3 and reported in Table~2 of the main paper.

\subsection{Distillation and Hyperparameters}
\label{appendix: model architecture}

When training AFO Student, we employ a dynamic distillation weight:
$\max(0,\, 1 - \text{student rate}/\text{teacher rate})$,
which decreases as the student catches up to the teacher~\cite{agarwal2022reincarnatingr}. 
Here, student rate and teacher rate denote their success rates. The weight reaches zero once the student matches or exceeds the teacher, after which training reduces to standard BC.
Trajectories containing VFM errors are retained without filtering or correction. The student's action supervision always comes from expert ground-truth actions, while the teacher provides only feature-level supervision.
Table~\ref{table:hyperparams} summarizes the training hyperparameters used across all experiments.

\vspace{0.2cm}

\begin{table}[h]
\caption{\textbf{Training hyperparameters.}}
\label{table:hyperparams}
\vspace{-0.1cm}
\centering
\resizebox{\linewidth}{!}{%
\begin{tabular}{lc}
\toprule
\textbf{Hyperparameter} & \textbf{Value} \\
\midrule
Optimizer & Adam \\
Learning rate & $1\!\times\!10^{-3}$ \\
Momentum & $\beta_1,\beta_2 = 0.9,0.999$ \\
Batch size & 256 \\
LR schedule & cosine ($10^{-3}{\to}10^{-6}$) \\
Warmup epochs & 5 \\
Training epochs & \begin{tabular}{c}Sim:100, Real:1000 (Sweep, Pick), \\
1500 (Toolbox)\end{tabular} \\
Data augmentation & Random shift (pad=12) \\
History window & 3 frames \\
BC loss & MSE \\
Distillation loss & Cosine similarity \\
Distillation weight & $1.0 \to 0.0$ (dynamic) \\
\bottomrule
\end{tabular}
}
\end{table}

\subsection{Molmo Prompts for SAM2}
\label{appendix: molmo prompts}

SAM2 requires spatial prompts for the first frame of each episode.
We use Molmo~\cite{deitke2025molmo} to generate these prompts automatically.
Below are the per-task natural-language queries used across all experiments.
For tasks with small objects (e.g., \textit{Pen}), we upscale input frames
from $256{\times}256$ to $512{\times}512$ before querying Molmo, then
downscale the returned coordinates.

\begin{itemize}
  \item \textbf{Relocate}
  \begin{itemize}
    \item green ball: \textit{Point to 1 location of the green ball.}
    \item blue ball: \textit{Point to 1 location of the blue ball.}
  \end{itemize}
  \item \textbf{Pen}
  \begin{itemize}
    \item blue pen – side: \textit{Point to 2 locations of both ends of the blue pen placed on the hand.}
    \item blue pen – center: \textit{Point to 1 location of the center of the blue pen placed on the hand.}
    \item green pen – side: \textit{Point to 2 locations of both ends of the green pen.}
    \item green pen – center: \textit{Point to 1 location of the center of the green pen.}
  \end{itemize}
  \item \textbf{Assembly}
  \begin{itemize}
    \item red object: \textit{Point to 1 location of the red cylinder.}
    \item gray ring: \textit{Point to 1 location of the gray circular ring.}
    \item green handle: \textit{Point to 1 location of the green rectangular block.}
  \end{itemize}
  \item \textbf{Button Press}
  \begin{itemize}
    \item red button: \textit{Point to 1 location of the red circle.}
    \item yellow box: \textit{Point to 1 location of the yellow box.}
  \end{itemize}
  \item \textbf{Drawer Open}
  \begin{itemize}
    \item handle: \textit{Point to 2 locations of the white handle.}
  \end{itemize}
  \item \textbf{Bin Picking}
  \begin{itemize}
    \item object: \textit{Point to 1 location of the light green cube.}
  \end{itemize}
  \item \textbf{Hammer}
  \begin{itemize}
    \item button: \textit{Point to 1 location of the gray button.}
    \item handle: \textit{Point to 1 location of the red handle of the hammer.}
    \item gray head – center: \textit{Point to 1 location of the center of the hammer's gray head.}
    \item gray head – top/bottom: \textit{Point to 2 locations of the top and bottom parts of the hammer's gray head.}
  \end{itemize}
\end{itemize}

While simpler prompts suffice for most tasks, tasks such as \textit{Pen}
require multi-point prompts to ensure consistent segmentation across
different trajectories.

\subsection{Latency Measurement Protocol} 
\label{supp:latency}

For the efficiency results reported in Table 1 and Table B3, we measure per-timestep policy inference latency with batch size 1 in FP32 on a single NVIDIA A100 GPU, following standard practice in ML inference benchmarking~\cite{mlperf}. Each measurement feeds observations through the full pipeline including image preprocessing, and we report the mean over repeated timed iterations with GPU synchronization around each iteration. The hardware is kept fixed across all compared models, and we report latency together with parameter count~\cite{efficiencyMisnomer}. For AFO (Teacher), we report the episode-averaged per-timestep latency, computed by summing the inference time over all timesteps from \(t=0\) to \(T\) and dividing by the total number of timesteps. At \(t=0\), the pipeline runs Molmo prompt generation, SAM 2 initialization, and DAM 2 inference. At subsequent timesteps, it runs DAM 2 inference and SAM 2 video tracking. Thus, the one-time initialization cost is included and amortized over the episode length.

\subsection{Annotation Criteria for Fig. 4 (d)}
For the analysis in Fig. 4 (d), we collect 150 teacher rollouts on Adroit Relocate
and assign each rollout a binary label indicating whether both
SAM 2 and DAM 2 remain accurate for every timestep of the episode. A rollout is labeled as
containing a VFM error if at least one misprediction occurs at any timestep. A segmentation
misprediction is a frame in which SAM 2 fails to cover the target object or assigns the mask to an
irrelevant region such as the robot hand or the table. A depth misprediction is a frame in which
DAM 2 places the target object at a clearly incorrect relative distance, for example by rendering a
shadow as a nearby surface. Labels are assigned by 2 human annotators inspecting every frame of each rollout.

\subsection{Statistical Significance}
\label{supp:significance}

For each method we compute the per-seed average success rate within each task group, giving five
  independent samples per method, and compare AFO (Student) against every other method with a
  two-sided Welch's $t$-test~\cite{welchTest}, which does not assume equal variances. Table~\ref{tab:significance}
  reports the resulting $p$-values on Adroit, on MetaWorld, and on the 7-task average.
  AFO (Student) improves significantly over every baseline and over its own teacher on the 7-task
  average, and on Adroit every comparison is significant. 
  On MetaWorld, only
  three of the seven comparisons reach significance, as this suite is close to ceiling for every
  method. Two of the five tasks are solved at or near $100\%$ by all methods, and the remaining
  margins are often smaller than the four-percentage-point resolution of a 25-rollout evaluation,
  for example $+0.64$ over R3M. 

\begin{table*}[t]
\centering
\caption{Significance of AFO (Student) over each method, two-sided Welch's $t$-test over five seeds.}
\label{tab:significance}
\resizebox{0.8\linewidth}{!}{%
\begin{tabular}{l|ccc|ccc}
\toprule
\multirow{2}{*}[-0.1cm]{Method} & \multicolumn{3}{c|}{Success Rate} & \multicolumn{3}{c}{$p$-value} \\
\cmidrule(lr){2-4} \cmidrule(lr){5-7}
 & \begin{tabular}{c}Adroit\\(2 tasks)\end{tabular} & \begin{tabular}{c}MetaWorld\\(5 tasks)\end{tabular} & \begin{tabular}{c}Average\\(7 tasks)\end{tabular} & \begin{tabular}{c}Adroit\\(2 tasks)\end{tabular} & \begin{tabular}{c}MetaWorld\\(5 tasks)\end{tabular} & \begin{tabular}{c}Average\\(7 tasks)\end{tabular}  \\
\drule
DINOv2~\cite{oquab2023dinov2}        & 66.00\std{3.16} & 95.52\std{1.45} & 87.09\std{1.60} & $<$0.0001 & 0.1668 & 0.0005 \\
R3M~\cite{nair2022r3m} & 71.20\std{3.63} & 96.32\std{1.21} & 89.14\std{0.57} & 0.0014 & 0.4871 & 0.0048 \\
MVP~\cite{radosavovic2022mvp}           & 62.40\std{5.55} & 93.76\std{1.73} & 84.80\std{0.96} & 0.0009 & 0.0152 & $<$0.0001 \\
VC-1~\cite{majumdar2023vc1}          & 60.80\std{4.15} & 93.76\std{1.64} & 84.34\std{0.87} & 0.0001 & 0.0130 & $<$0.0001 \\
Theia~\cite{shang2024theia}         & 68.80\std{5.40} & 94.88\std{2.37} & 87.43\std{2.74} & 0.0045 & 0.1450 & 0.0097 \\
AFO (Teacher) & 72.40\std{2.19} & 92.96\std{2.49} & 87.09\std{1.54} & 0.0001 & 0.0196 & 0.0005 \\
\midrule
LfS~\cite{hansen2022lfs} & 61.60\std{3.58} & 93.76\std{2.85} & 84.57\std{2.02} & $<$0.0001 & 0.0682 & 0.0001 \\
AFO (Student) & 81.60\std{1.67} & 96.96\std{1.54} & 92.57\std{1.51} & --- & --- & --- \\
\bottomrule
\end{tabular}
}
\end{table*}

%%%%%%%%%%%%%%%%%%%%%%%%%%%%%%%%%%%%%%%%%%%%%%%%%%%%%%%%%%%%
\section{Full Performance Table}
\label{supp:full_results}

%%%%%%%%%%%%%%%%%%%%%%%%%%%%%%%%%%%%%%%%%%%%%%%%%%%%%%%%%%%%

Table~\ref{Appendix table: full performance table} reports complete per-task
results with standard deviations across five seeds for all models evaluated in
the main paper (Table 1). 
Table~\ref{tab:ablation_pertask} provides the per-task results for main Table 5, and Table~\ref{tab:theia_pertask} provides those for main Table 4.

  \begin{table*}[t]
  \centering
  \caption{\textbf{Full performance table with standard deviations.} We report the mean success rate and standard deviation ($\pm$) across 5 seeds. For all baselines, random-shift data augmentation is applied. For ViT baselines, spatial tokens (not \texttt{[CLS]}) are used as policy input.}
  \label{Appendix table: full performance table}
  \resizebox{\textwidth}{!}{%
  \begin{tabular}{l|cc|cc|ccccc|ccc}
  \toprule
  \multirow{2}{*}[-0.1cm]{Model} & \multicolumn{2}{c|}{Efficiency} & \multicolumn{2}{c|}{Adroit} & \multicolumn{5}{c|}{MetaWorld} & \multicolumn{3}{c}{Average} \\
  \cmidrule(lr){2-3} \cmidrule(lr){4-5} \cmidrule(lr){6-10} \cmidrule(lr){11-13}
   & Inference & \#Params & Pen & Relocate & Assembly & Bin & Button & Drawer & Hammer & Adroit & MetaWorld & 7 tasks \\
  \drule
  DINOv2~\cite{oquab2023dinov2} & 12.48ms & 88.26M & 72.80\std{3.35} & 59.20\std{9.12} & \textbf{100.00}\std{0.00} & 89.60\std{6.69} & 89.60\std{4.56} & \textbf{100.00}\std{0.00} & 98.40\std{2.19} & 66.00\std{3.16} & 95.52\std{1.45} & 87.09\std{1.60} \\
  R3M~\cite{nair2022r3m} & 6.41ms & 25.60M & 72.00\std{7.48} & 70.40\std{3.58} & \textbf{100.00}\std{0.00} & 88.80\std{5.22} & \textbf{96.80}\std{3.35} & \textbf{100.00}\std{0.00} & 96.00\std{0.00} & 71.20\std{3.63} & 96.32\std{1.21} & \underline{89.14}\std{0.57} \\
  MVP~\cite{radosavovic2022mvp} & 7.11ms & 87.40M & 68.00\std{4.90} & 56.80\std{6.57} & \textbf{100.00}\std{0.00} & 95.20\std{4.38} & 75.20\std{10.73} & \textbf{100.00}\std{0.00} & 98.40\std{2.19} & 62.40\std{5.55} & 93.76\std{1.73} & 84.80\std{0.96} \\
  VC-1~\cite{majumdar2023vc1} & 6.84ms & 87.40M & 68.80\std{3.35} & 52.80\std{6.57} & \textbf{100.00}\std{0.00} & 93.60\std{3.58} & 75.20\std{7.69} & \textbf{100.00}\std{0.00} & \textbf{100.00}\std{0.00} & 60.80\std{4.15} & 93.76\std{1.64} & 84.34\std{0.87} \\
  Theia~\cite{shang2024theia} & 7.94ms & 88.40M & 72.00\std{4.90} & 65.60\std{7.80} & \textbf{100.00}\std{0.00} & 91.20\std{7.69} & 84.00\std{11.31} & \textbf{100.00}\std{0.00} & \underline{99.20}\std{1.79} & 68.80\std{5.40} & 94.88\std{2.37} & 87.43\std{2.74} \\
  \midrule
  CNN (LfS)~\cite{hansen2022lfs} & 2.11ms & 2.72M & 74.40\std{7.27} & 48.80\std{4.38} & 76.00\std{15.75} & 96.80\std{3.35} & \underline{96.00}\std{4.90} & \textbf{100.00}\std{0.00} & \textbf{100.00}\std{0.00} & 61.60\std{3.58} & 93.76\std{2.85} & 84.57\std{2.02} \\
  CNN + DAM & 42.78ms & 100.34M & \underline{77.60}\std{6.07} & 56.80\std{3.35} & 88.80\std{7.69} & \textbf{99.20}\std{1.79} & \underline{96.00}\std{4.00} & \textbf{100.00}\std{0.00} & \textbf{100.00}\std{0.00} & 67.20\std{3.63} & \underline{96.80}\std{1.13} & 88.34\std{1.65} \\
  CNN + SAM & 70.95ms & 8.10B & 75.20\std{5.22} & 72.00\std{7.48} & 64.80\std{18.20} & 94.40\std{4.56} & 92.00\std{8.00} & \textbf{100.00}\std{0.00} & \underline{99.20}\std{1.79} & \underline{73.60}\std{4.77} & 90.08\std{5.32} & 85.37\std{4.51} \\
  AFO (All): CNN + DAM + SAM & 110.07ms & 8.20B & 69.60\std{4.56} & \underline{75.20}\std{3.35} & 75.20\std{10.35} & \underline{98.40}\std{2.19} & 92.00\std{6.32} & \underline{99.20}\std{1.79} & \textbf{100.00}\std{0.00} & 72.40\std{2.19} & 92.96\std{2.49} & 87.09\std{1.54} \\
  \textcolor{gray}{AFO (Teacher w/ GT)} & \textcolor{gray}{3.84ms} & \textcolor{gray}{3.04M} & \textcolor{gray}{76.00\std{4.00}} & \textcolor{gray}{96.80\std{3.35}} & \textcolor{gray}{94.40\std{2.19}} & \textcolor{gray}{94.40\std{5.37}} & \textcolor{gray}{99.20\std{1.79}} & \textcolor{gray}{100.00\std{0.00}} & \textcolor{gray}{100.00\std{0.00}} & \textcolor{gray}{86.40\std{2.97}} & \textcolor{gray}{97.60\std{0.80}} & \textcolor{gray}{94.40\std{1.30}} \\
  \midrule
  AFO (For-One): CNN distilled from All & 2.11ms & 2.72M & \textbf{79.20}\std{3.35} & \textbf{84.00}\std{2.83} & \underline{92.00}\std{4.90} & 96.80\std{1.79} & \underline{96.00}\std{4.90} & \textbf{100.00}\std{0.00} & \textbf{100.00}\std{0.00} & \textbf{81.60}\std{1.67} & \textbf{96.96}\std{1.54} & \textbf{92.57}\std{1.51} \\
  \bottomrule
  \end{tabular}}
  \end{table*}

\begin{table*}[t]
\centering
\caption{\textbf{Per-task results for the comparison with the Theia distillation objective.} Mean success rate and standard deviation ($\pm$) across 5 seeds.}
\label{tab:theia_pertask}
\resizebox{\textwidth}{!}{%
\begin{tabular}{l|ccc|cc|ccccc|ccc}
\toprule
\multirow{2}{*}[-0.1cm]{Method} & \multicolumn{3}{c|}{Distillation Configuration} & \multicolumn{2}{c|}{Adroit} & \multicolumn{5}{c|}{MetaWorld} & \multicolumn{3}{c}{Average} \\
\cmidrule(lr){2-4} \cmidrule(lr){5-6} \cmidrule(lr){7-11} \cmidrule(lr){12-14}
 & Dataset & Distill Target & Capacity & Pen & Relocate & Assembly & Bin & Button & Drawer & Hammer & Adroit & MetaWorld & 7 tasks \\
\drule
Theia (Orig.)~\cite{shang2024theia} & \cellcolor{gray!20}\emph{ImageNet} & \cellcolor{gray!20}\emph{VFM Features} & \cellcolor{gray!20}\emph{ViT} & 72.00\std{4.90} & 65.60\std{7.80} & \textbf{100.00}\std{0.00} & 91.20\std{7.69} & 84.00\std{11.31} & \textbf{100.00}\std{0.00} & \underline{99.20}\std{1.79} & 68.80\std{5.40} & 94.88\std{2.37} & 87.43\std{2.74} \\
 & Downstream & \cellcolor{gray!20}\emph{VFM Features} & \cellcolor{gray!20}\emph{ViT} & \textbf{82.40}\std{3.58} & 79.20\std{9.55} & \textbf{100.00}\std{0.00} & 92.00\std{6.32} & 89.60\std{5.37} & \textbf{100.00}\std{0.00} & 92.80\std{5.93} & 80.80\std{5.02} & 94.88\std{2.09} & 90.86\std{1.40} \\
 & Downstream & VFM Outputs & \cellcolor{gray!20}\emph{ViT} & \underline{80.80}\std{3.35} & \textbf{96.80}\std{3.35} & \textbf{100.00}\std{0.00} & \textbf{100.00}\std{0.00} & \textbf{99.20}\std{1.79} & \textbf{100.00}\std{0.00} & \underline{99.20}\std{1.79} & \textbf{88.80}\std{3.03} & \textbf{99.68}\std{0.44} & \textbf{96.57}\std{0.81} \\
AFO (Student) & Downstream & VFM Outputs & CNN & 79.20\std{3.35} & \underline{84.00}\std{2.83} & \underline{92.00}\std{4.90} & \underline{96.80}\std{1.79} & \underline{96.00}\std{4.90} & \textbf{100.00}\std{0.00} & \textbf{100.00}\std{0.00} & \underline{81.60}\std{1.67} & \underline{96.96}\std{1.54} & \underline{92.57}\std{1.51} \\
\bottomrule
\end{tabular}}
\end{table*}

\begin{table*}[t]
\centering
\caption{\textbf{Per-task results for the ablation study.} Mean success rate and standard deviation ($\pm$) across 5 seeds.}
\label{tab:ablation_pertask}
\resizebox{\textwidth}{!}{%
\begin{tabular}{ccc|cc|cc|ccccc|ccc}
\toprule
\multicolumn{3}{c|}{Distillation Configuration} & \multicolumn{2}{c|}{Teacher} & \multicolumn{2}{c|}{Adroit} & \multicolumn{5}{c|}{MetaWorld} & \multicolumn{3}{c}{Average} \\
\cmidrule(lr){1-3} \cmidrule(lr){4-5} \cmidrule(lr){6-7} \cmidrule(lr){8-12} \cmidrule(lr){13-15}
Loss & Alignment & $\lambda$ & Segment. & Depth & Pen & Relocate & Assembly & Bin & Button & Drawer & Hammer & Adroit & MetaWorld & 7 tasks \\
\drule
\multicolumn{5}{c|}{\emph{BC-only from scratch (LfS), no distillation}} & 74.40\std{7.27} & 48.80\std{4.38} & 76.00\std{15.75} & 96.80\std{3.35} & \underline{96.00}\std{4.90} & \textbf{100.00}\std{0.00} & \textbf{100.00}\std{0.00} & 61.60\std{3.58} & 93.76\std{2.85} & 84.57\std{2.02} \\
\midrule
\cellcolor{gray!20}$\ell_2$ & per-modality & dynamic & SAM & DAM & \textbf{79.20}\std{7.69} & \underline{85.60}\std{2.19} & 82.40\std{9.21} & \textbf{100.00}\std{0.00} & \textbf{96.80}\std{5.22} & \textbf{100.00}\std{0.00} & \textbf{100.00}\std{0.00} & \textbf{82.40}\std{2.97} & 95.84\std{2.49} & \underline{92.00}\std{2.49} \\
cos & \cellcolor{gray!20}\emph{fused} & dynamic & SAM & DAM & 74.40\std{5.37} & 83.20\std{3.35} & \underline{91.20}\std{7.16} & \underline{99.20}\std{1.79} & \underline{96.00}\std{2.83} & \textbf{100.00}\std{0.00} & \textbf{100.00}\std{0.00} & 78.80\std{3.63} & \textbf{97.28}\std{1.56} & \underline{92.00}\std{1.46} \\
cos & per-modality & \cellcolor{gray!20}\emph{fixed} 1.0 & SAM & DAM & \underline{76.00}\std{2.83} & 82.40\std{8.29} & 87.20\std{9.55} & 98.40\std{2.19} & \underline{96.00}\std{4.90} & \textbf{100.00}\std{0.00} & \textbf{100.00}\std{0.00} & 79.20\std{5.40} & 96.32\std{2.16} & 91.43\std{2.10} \\
\midrule
cos & per-modality & dynamic & \cellcolor{gray!20}-- & DAM & 74.40\std{6.07} & 68.00\std{9.38} & 88.00\std{9.38} & 98.40\std{2.19} & \underline{96.00}\std{4.00} & \textbf{100.00}\std{0.00} & \textbf{100.00}\std{0.00} & 71.20\std{6.72} & 96.48\std{2.37} & 89.26\std{3.49} \\
cos & per-modality & dynamic & SAM & \cellcolor{gray!20}-- & \underline{76.00}\std{4.90} & 83.20\std{7.16} & 88.80\std{7.16} & \underline{99.20}\std{1.79} & 93.60\std{5.37} & \textbf{100.00}\std{0.00} & \textbf{100.00}\std{0.00} & 79.60\std{4.34} & 96.32\std{1.34} & 91.54\std{1.48} \\
\midrule
cos & per-modality & dynamic & SAM & DAM & \textbf{79.20}\std{3.35} & 84.00\std{2.83} & \textbf{92.00}\std{4.90} & 96.80\std{1.79} & \underline{96.00}\std{4.90} & \textbf{100.00}\std{0.00} & \textbf{100.00}\std{0.00} & \underline{81.60}\std{1.67} & \underline{96.96}\std{1.54} & \textbf{92.57}\std{1.51} \\
\bottomrule
\end{tabular}}
\end{table*}

%%%%%%%%%%%%%%%%%%%%%%%%%%%%%%%%%%%%%%%%%%%%%%%%%%%%%%%%%%%%
\section{Qualitative VFM Outputs}
\label{supp:qualitative}
%%%%%%%%%%%%%%%%%%%%%%%%%%%%%%%%%%%%%%%%%%%%%%%%%%%%%%%%%%%%

\begin{figure*}[ht]
\begin{center}
\includegraphics[width=\textwidth]{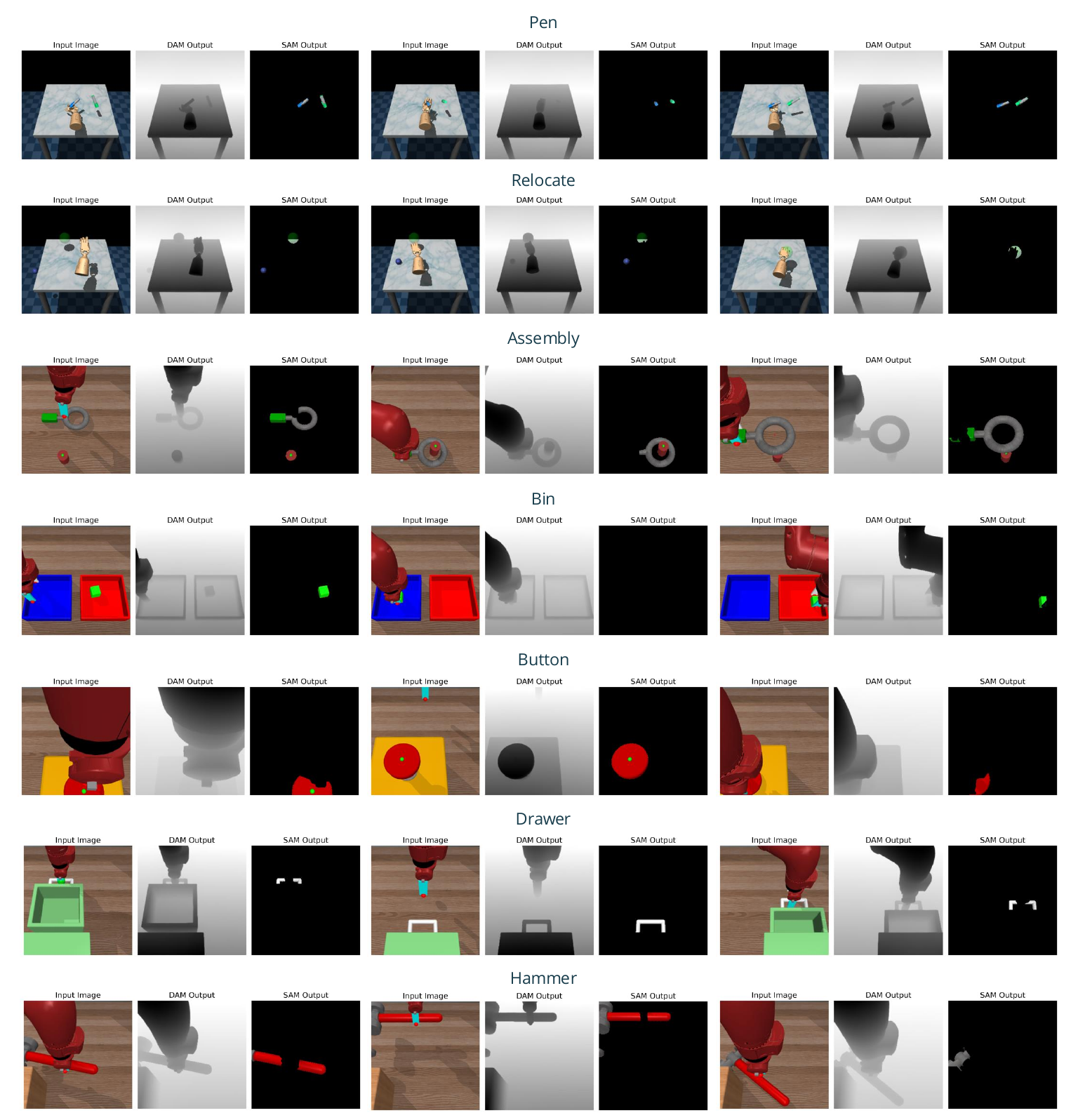}
\end{center}
\vspace{-2mm}
\caption{\textbf{Qualitative outputs of Depth Anything v2 and Segment Anything v2 across tasks.}
Each row shows (from left) the RGB input, depth map (normalized to $[0,1]$),
and segmentation mask overlay.
Both successful predictions and failure cases are illustrated.}
\label{appendix_fig:additional_qualitative}
\end{figure*}

Figure~\ref{appendix_fig:additional_qualitative} visualizes the outputs of
Depth Anything v2~\cite{depth_anything_v2} and Segment Anything v2~\cite{ravi2024sam2} across all Adroit~\cite{rajeswaran2018adroit} and MetaWorld~\cite{yu2020metaworld} tasks used in our experiments.
Depth maps are linearly normalized to $[0, 1]$.
Segmentation results are shown as masked overlays on the original RGB image.
While VFMs succeed in the majority of cases, occasional failures occur, \eg, failed segmentation of the hammer handle (last row).
This motivates our distillation approach that eliminates VFM dependency at inference time.

\clearpage

{
\small
\bibliographystyle{ieeenat_fullname}
\bibliography{IEEEabrv,reference}
}